\PassOptionsToPackage{table}{xcolor}
\documentclass{article} 
\usepackage{iclr2027_conference,times}

\usepackage{amsmath,amsfonts,bm}

\def\eqref#1{equation~\ref{#1}}

\def\1{\bm{1}}

\DeclareMathAlphabet{\mathsfit}{\encodingdefault}{\sfdefault}{m}{sl}
\SetMathAlphabet{\mathsfit}{bold}{\encodingdefault}{\sfdefault}{bx}{n}

\newcommand{\KL}{D_{\mathrm{KL}}}

\usepackage{algorithm}
\usepackage{algpseudocode}
\usepackage{hyperref}
\usepackage{xurl}
\usepackage{booktabs}
\usepackage{amsmath}
\usepackage{amssymb}
\usepackage{graphicx}
\usepackage{wrapfig}
\usepackage{needspace}
\usepackage{flafter}
\usepackage{placeins}
\usepackage{xcolor}
\definecolor{scoregain}{RGB}{0,112,55}
\definecolor{scoreloss}{RGB}{175,48,35}
\definecolor{methodshade}{RGB}{244,244,244}
\definecolor{oursshade}{RGB}{225,244,249}
\hypersetup{colorlinks=false,pdfborder={0 0 1},citebordercolor={0 1 0},linkbordercolor={1 0 0},urlbordercolor={0 1 1}}
\definecolor{linkpurple}{RGB}{88,92,168}
\newcommand{\plainurl}[1]{{\hypersetup{pdfborder={0 0 0}}\urlstyle{same}\textcolor{linkpurple}{\url{#1}}}}

\newcommand{\mixweight}{\lambda}
\newcommand{\resscale}{\gamma}

\title{From Lookups to Lessons: Turning Retrieved Knowledge into Parametric Memory}

\author{
\parbox{\textwidth}{\centering
Changjiang Han$^{1}$, Bowei He$^{1,2}$, Ye Yuan$^{2,3}$, Weixu Zhang$^{2,3}$, \\
Haolun Wu$^{2,3}$, Xue Liu$^{1,2,3}$, Jikun Kang$^{4,*}$ \\
{\normalfont $^{1}$MBZUAI, $^{2}$McGill University, $^{3}$Mila, $^{4}$Salesforce} \\
{\normalfont\fontfamily{zi4}\selectfont Changjiang.Han@mbzuai.ac.ae}}
}

\iclrfinalcopy
\begin{document}

\maketitle
\lhead{Preprint}
\begingroup
\renewcommand{\thefootnote}{}
\makeatletter
\long\def\@makefntext#1{\noindent #1}
\makeatother
\footnotetext{$^{*}$Corresponding author.}
\addtocounter{footnote}{-1}
\endgroup

\begin{abstract}
Retrieval-augmented generation (RAG) helps large language models (LLMs) answer factual questions beyond their training data, but it adds storage and inference costs.
Distilling previously retrieved information into parametric memories enables reuse without online retrieval.
However, training the memory to mimic a retriever rewards accurate independent predictions, which do not necessarily improve the final output quality when combined with the LLM's existing predictions.
To address this, we introduce \textbf{ResMem}, in which the LLM reads retrieved text offline to generate teacher targets.
This design avoids storing massive token representations, 
unlike existing parametric memory approaches such as MLP Memory, which relies on k-nearest neighbor (kNN) datastores. 
Furthermore, by using the LLM as an offline teacher that reads privileged retrieved evidence alongside the input, we train a parametric memory to learn an additive logit residual.
This residual aligns the predictions of a student sharing the same backbone with the teacher distribution, while self-distillation updates remain confined to the parametric memory.
On a Wikipedia subset, ResMem reduces the storage retained for teacher construction by 94.7\% compared with the evaluated kNN pipeline. 
Across five question answering datasets, ResMem raises mean token F1 from 32.83\% for the frozen LLM to 34.48\%.
It also breaks 21\% fewer of the LLM's correct answers than a kNN-trained memory.
These findings validate our motivation of treating retrieval as a learning resource, with parametric memory shaped by how retrieved knowledge is used to inform the model's future predictions.
Code and model weights are available at \plainurl{https://github.com/SSRand/From-Lookups-to-Lessons} and \plainurl{https://huggingface.co/RandHan/ResMem-Mistral-7B-v0.3}.
\end{abstract}

\section{Introduction}
\label{sec:intro}

Nowadays, many LLM applications need information beyond the current prompt, such as knowledge from external corpora and facts introduced in earlier interactions \citep{park2023generativeagents,zhang2025memorysurvey}.
One option is to keep such information in the model's active context.
However, longer context windows do not ensure reliable use of information at every position \citep{liu2024lostmiddle}, and growing histories still require managing a finite active context \citep{packer2024memgptllmsoperatingsystems}.
Retrieval-augmented generation (RAG) is a widely used way to recover relevant evidence from external storage \citep{lewis2020rag}.
This access requires maintaining a corpus and its index, while retrieving and reading the evidence adds inference latency.
At corpus scale, these costs motivate storing knowledge for use without online retrieval.

\begin{figure}[t]
\centering
\includegraphics[width=0.88\linewidth]{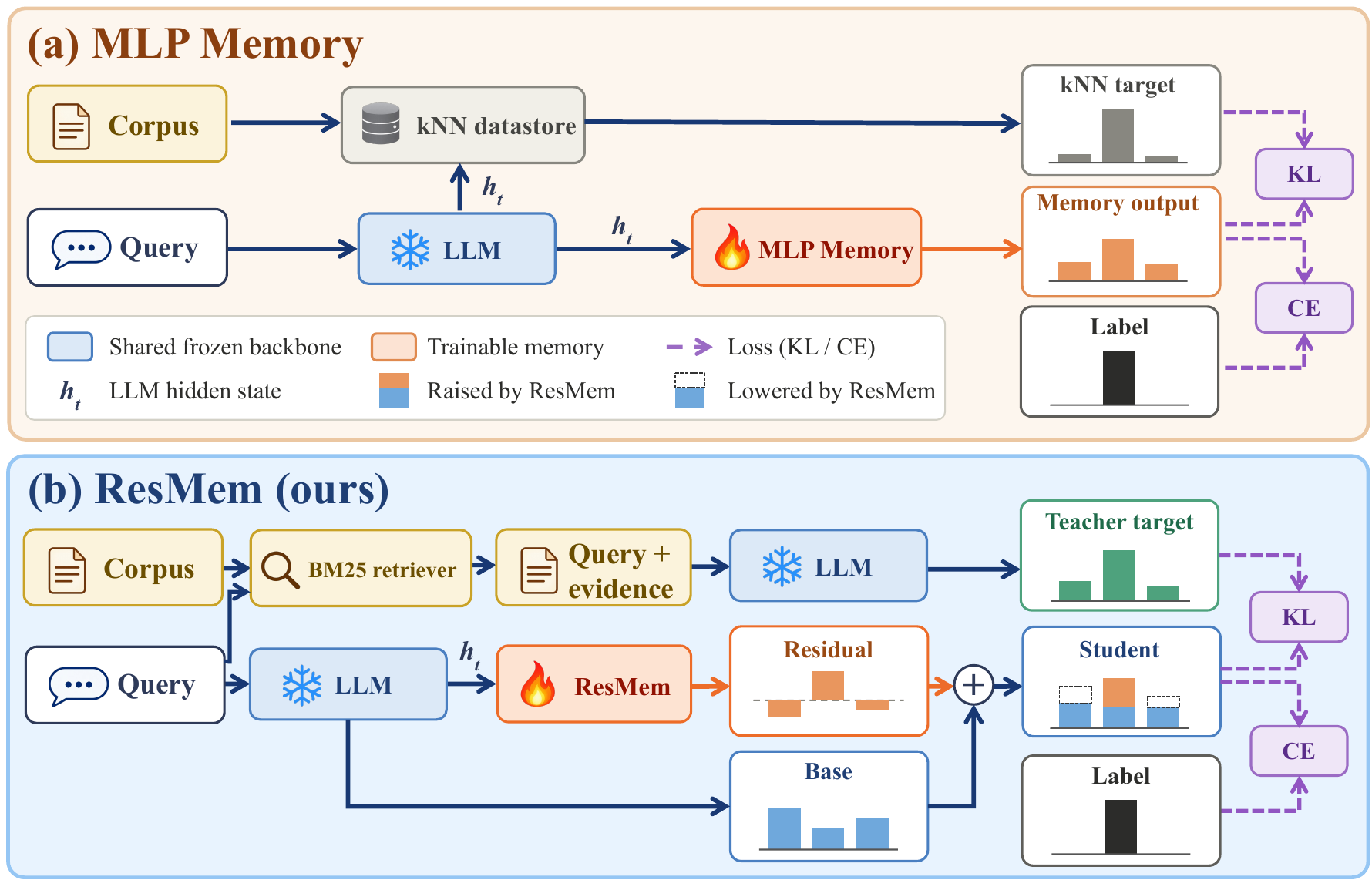}
\caption{Training pipelines.
(a) MLP Memory fits a memory alone to kNN targets; the base enters only at inference (Eq.~\ref{eq:mlp-readout}).
(b) ResMem fits a logit residual so that base plus residual matches an evidence-reading teacher; inference drops the teacher and scales the residual by $\resscale$ (Eq.~\ref{eq:inference}).}
\label{fig:framework}
\vspace{-5mm}
\end{figure}

Retriever-pretrained parametric memories offer such retrieval-free storage: they learn what a retriever would return and keep it in network weights, so inference needs no index or search.
Building on kNN language models \citep{khandelwal2020knnlm}, MLP Memory and Memory Decoder train a memory to reproduce a retriever's next-token distribution, then interpolate its prediction with a frozen base model \citep{wei2025mlpmemory,cao2025memorydecoder}.
MLP Memory reports storage compression and faster inference, while Memory Decoder enables reuse across compatible base models.
Both memories, however, learn to match the retriever alone and meet the base model only at inference.
Their training loss never sees the base prediction (Figure~\ref{fig:framework}a).
A memory that reproduces the retriever well can therefore still change the combined prediction in unhelpful ways.
For example, it can override a correct base answer.

We introduce ResMem, which trains the memory through the base prediction it will later modify (Figure~\ref{fig:framework}b).
The frozen base model serves as its own teacher by reading the query together with retrieved evidence, while the student sees only the query.
The memory adds a logit residual to the student's logits, and the distillation and next-token losses are computed on this sum.
Because the loss is computed on the base prediction plus the residual, positions where the base already matches the training target give little gradient.
The memory thus learns a correction to the base, not a second copy of the retriever.
This builds on context distillation and retrieval-free dialogue distillation \citep{snell2022context,zhu2024ra2fd}.
Inference then needs only the frozen base and the trained memory, with a coefficient $\resscale$ setting how strongly the residual adjusts the base (Eq.~\ref{eq:inference}).

Because retriever-pretrained memories draw their supervision from a kNN datastore of corpus representations, they also pay a construction cost.
Memory Decoder names the neighbor search over this datastore as a training bottleneck \citep[Sec.~9]{cao2025memorydecoder}, and later work scales the search with compression and sharding \citep{wei2026memorydecoderscale}.
An evidence-reading teacher needs no such datastore.
ResMem retrieves passages with BM25 and caches only the teacher's top-64 targets, which cuts retained construction storage from 602.0 to 32.0 GiB on a subset of English Wikipedia, a 94.7\% ($18.8\times$) reduction (Section~\ref{sec:storage}).

On five knowledge-QA datasets, ResMem improves exact match (EM) and token F1 over the base model on every dataset and raises mean F1 from 32.83 to 34.48.
Compared with MLP Memory, it corrects 11\% more base errors and breaks 21\% fewer correct base answers, so most of its advantage comes from avoided harms.
Under a shared teacher, the residual objective mainly preserves correct answers, and its extra repairs come with the evidence-reading teacher.
Memory Decoder falls below both memories under the same teacher.

Our main contributions are: (i)~We introduce residual self-distillation, which trains an additive logit residual through the joint base-plus-memory prediction and keeps it at inference.
(ii)~We replace the kNN datastore with text retrieval and an evidence-reading teacher.
This cuts retained construction storage by 94.7\% ($18.8\times$).
The teacher's targets also keep the shape of the base prediction.
The kNN targets are much sharper.
(iii)~We evaluate the recipe and its parts on knowledge QA, answer-level corrections and harms, and shared-teacher ablations.

\section{Related Work}
\label{sec:preliminary}

\textbf{Retrieval and parametric memory.} Retrieval keeps information outside model parameters.
RAG, RETRO, and Atlas condition generation on retrieved text \citep{lewis2020rag,borgeaud2022retro,izacard2022atlas}, and xRAG compresses each retrieved document into a one-token representation \citep{cheng2024xrag}. kNN-LM instead forms a next-token distribution from neighboring contexts \citep{khandelwal2020knnlm}.
Memorizing Transformers and LongMem attend to stored representations of earlier tokens \citep{wu2022memorizing,wang2023longmem}.
Product quantization, GPU similarity search, and efficient kNN-LMs reduce the cost of such stores \citep{jegou2011pq,johnson2021faiss,he2021efficientknn}.
All of these methods query a store during inference.
Parametric memories instead learn to reproduce a store, building on the view of feed-forward layers as key--value memories \citep{geva2021ffn}. kNN knowledge distillation moves retrieval into preprocessing for translation \citep{yang2022knnkd}.
MLP Memory and Memory Decoder train separate parametric memories to imitate retriever distributions \citep{wei2025mlpmemory,cao2025memorydecoder}.
Memory Decoder at Scale extends this route to larger memories and corpora with compressed, sharded search \citep{wei2026memorydecoderscale}.
MemSFT adds learned token-level routing between a retrieval-supervised memory and the frozen backbone \citep{wang2026memsft}.
We instead train the memory's contribution through the joint prediction.
Parametric RAG still retrieves documents, each encoded as LoRA parameters \citep{su2025parametricrag}.
Test-time training layers, Titans, and $\delta$-mem update memory states online while processing a stream \citep{sun2025ttt,behrouz2025titans,lei2026deltamem}.
ResMem instead trains its memory offline.

\textbf{Distillation and output composition.} Knowledge distillation transfers a teacher's predictions to a student \citep{hinton2015distilling}, and the teacher may use privileged information that is absent at deployment \citep{lopezpaz2016privileged}.
Context distillation internalizes the benefit of extra instructions, examples, or reasoning in language models \citep{snell2022context}.
Cartridges uses it to learn compact KV caches offline \citep{eyuboglu2025cartridges}.
TTCD updates a short-context student's fast weights online from long-context supervision \citep{wang2026ttcd}.
Born-Again Networks show that the teacher need not be larger than the student \citep{furlanello2018bornagain}.
In ResMem, teacher and student share the frozen backbone, and only the parametric memory is trained.
Retrieval-conditioned distillation carries the benefits of retrieval into retrieval-free students, as in QUILL, RA2FD, and InstructRetro \citep{srinivasan2022quill,zhu2024ra2fd,wang2024instructretro}.
TS-Memory distills kNN-induced corrections into a time-series adapter \citep{lyu2026tsmemory}.
TransMem is the closest work.
It also trains a memory by evidence-conditioned self-distillation with a shared frozen backbone \citep{lei2026transmem}.
However, its student reads the full long context and outputs gated hidden-state shifts.
Our student lacks the evidence and outputs a logit residual.
DExperts, contrastive decoding, proxy-tuning, and context-aware decoding combine or contrast model predictions at decoding time \citep{liu2021dexperts,li2023contrastive,liu2024proxy,shi2024cad}.
ResMem instead learns its residual through a loss on the combined prediction and needs evidence only offline.
ResMem also trains on fixed teacher-forced continuations.
On-policy distillation, OPSD, and OPSDL use student-generated trajectories instead \citep{agarwal2024onpolicy,zhao2026opsd,zhang2026opsdl}.

\section{Methodology}
\label{sec:method}

ResMem writes external knowledge into a parametric memory while keeping the language model frozen.
The memory architecture follows MLP Memory.
ResMem changes the teacher and the objective (Section~\ref{sec:loss}), and the teacher's cached targets replace the kNN datastore (Section~\ref{sec:storage-cost}).

\subsection{Memory input and architecture}
For a token sequence $x$, causal position $t$ predicts the next token $y_t=x_{t+1}$.
The frozen base produces logits $z_{B,t}\in\mathbb{R}^{|V|}$ over vocabulary $V$.
At the same position, we extract $h_t\in\mathbb{R}^{d}$ at the input to the final Transformer block's feed-forward sublayer (Appendix~\ref{app:implementation}).

The memory maps this contextual vector to vocabulary logits, $s_{\theta,t}=m_\theta(h_t)\in\mathbb{R}^{|V|}$.
Following MLP Memory \citep{wei2025mlpmemory}, it consists of a stack of feed-forward blocks with RMS normalization and residual connections, followed by a final normalization and its own trainable vocabulary projection.
Each gated block acts on one position; the base representation already carries the preceding context, so the memory needs no attention over tokens.
Only memory parameters, including its output projection, are trained.

\subsection{Residual self-distillation}
\label{sec:loss}
\textbf{Distilling the privileged gain into memory.} On-policy self-distillation (OPSD) \citep{zhao2026opsd} turns privileged input into supervision: the same model, given information the student lacks, produces a dense token-level output distribution.
Teacher and student share one backbone, so this distribution lives in the base model's own output space.
It differs from the base prediction only through what the privileged input adds.
A residual on the base must learn exactly this part (Eq.~\ref{eq:residual-opt}).
ResMem uses retrieved evidence as the privileged input, as in context distillation \citep{lopezpaz2016privileged,snell2022context}, and distills this gain into the parametric memory.
The teacher $\pi_T(\cdot\mid c,x_{\le t})$ is the frozen model reading passages $c$ retrieved by BM25 \citep{robertson2009bm25} before the span $x$.
The student $\pi_S(\cdot\mid x_{\le t})=\mathrm{softmax}(z_{B,t}+s_{\theta,t})$ is the same model reading only $x$, plus the memory's residual.
Training minimizes the per-token divergence between the two roles over corpus prefixes,
\begin{equation}
\min_{\theta}\;\mathbb{E}_{x}\sum_{t}\KL\bigl(\pi_T(\cdot\mid c,x_{\le t})\,\big\|\,\pi_S(\cdot\mid x_{\le t})\bigr).
\label{eq:self-distill}
\end{equation}
Only the memory is updated.
The gain that the evidence gives the teacher is written into $\theta$, and the backbone stays fixed.

\textbf{Teacher targets.} Teacher and student follow the same continuation under teacher forcing.
After the evidence prefix, teacher outputs are aligned to the student's next-token targets.
The evidence changes the supervision, while the memory input remains $h_t$ from the student base.
Retrieval and teacher forwarding run once, offline.
At each supervised position, we cache the teacher's top-$K$ token probabilities with $K=64$.
We renormalize them and put zero mass elsewhere, which gives $p_{T,t}^{(K)}$.
This target is zero outside $K$ tokens, while $\pi_S$ has full support.
Eq.~\ref{eq:self-distill} uses the forward KL, because the reverse direction would be infinite.
Algorithm~\ref{alg:resmem} summarizes the procedure, and the paragraphs below specify the student and the loss.

\begin{algorithm}[t]
\caption{Residual self-distillation}
\label{alg:resmem}
\begin{algorithmic}[1]
\Require frozen language model $f$, corpus spans $\mathcal{X}$, retriever $R$, cache size $K$, loss weight $\alpha$
\ForAll{$x\in\mathcal{X}$}
  \State $c\gets R(x)$; \ $p_{T}^{(K)}\gets$ renormalized top-$K$ of $f(\cdot\mid[c;x])$ at every position
\EndFor
\Repeat
  \State sample spans $x\sim\mathcal{X}$; \ $(h_t,z_{B,t})\gets f(x)$
  \State $q_{R,t}\gets\mathrm{softmax}\bigl(z_{B,t}+m_\theta(h_t)\bigr)$
  \State $\mathcal{L}\gets\alpha\,\mathcal{L}_{\mathrm{KL}}+(1-\alpha)\,\mathcal{L}_{\mathrm{CE}}$
  \State $\theta\gets\theta-\eta\nabla_\theta\mathcal{L}$
\Until{the token budget is spent}
\State \Return $m_\theta$; inference uses $\mathrm{softmax}\bigl(z_{B,t}+\resscale\,m_\theta(h_t)\bigr)$
\end{algorithmic}
\end{algorithm}

\textbf{Joint prediction.} We train the memory through its joint prediction with the base ($\oplus$ in Figure~\ref{fig:framework}b), $z_{R,t}=z_{B,t}+s_{\theta,t}$ and $q_{R,t}=\mathrm{softmax}(z_{R,t})$.
During training the residual enters at unit scale, which corresponds to $\resscale=1$ in the inference rule of Section~\ref{sec:inference}.
Because the loss evaluates $q_{R,t}$ with $z_{B,t}$ fixed, the memory learns the change needed by the combined prediction.

\textbf{Loss components.} Let $\mathcal{C}$ contain all valid causal prediction positions in a training batch and $\mathcal{A}\subseteq\mathcal{C}$ the positions selected for label supervision.
As in knowledge distillation with soft targets and labels \citep{hinton2015distilling}, we combine teacher-to-student KL, $\mathcal{L}_{\mathrm{KL}}=|\mathcal{C}|^{-1}\sum_{t\in\mathcal{C}}\KL(p_{T,t}^{(K)}\,\|\,q_{R,t})$, with next-token cross-entropy, $\mathcal{L}_{\mathrm{CE}}=-|\mathcal{A}|^{-1}\sum_{t\in\mathcal{A}}\log q_{R,t}(y_t)$.
KL transfers the teacher's distribution over possible continuations; CE supervises the observed target token.
A hyperparameter $\alpha\in[0,1]$ weighs the two, $\mathcal{L}=\alpha\,\mathcal{L}_{\mathrm{KL}}+(1-\alpha)\,\mathcal{L}_{\mathrm{CE}}$.
Each term is averaged over its own selected positions, excluding padding.
In corpus-continuation training, both terms cover the complete valid span, so $\mathcal{A}=\mathcal{C}$.

\textbf{What the residual learns.} With $\mathcal{A}=\mathcal{C}$, each position contributes $\ell_t=-\sum_{v\in V}\tilde p_t(v)\log q_{R,t}(v)$ up to a constant, where the training target $\tilde p_t=\alpha\,p_{T,t}^{(K)}+(1-\alpha)\,e_{y_t}$ mixes the teacher with the one-hot label $e_{y_t}$.
The gradient of $\ell_t$ with respect to the residual logits is $\partial\ell_t/\partial s_{\theta,t}=\mathrm{softmax}(z_{B,t}+s_{\theta,t})-\tilde p_t$.
A position whose joint prediction already matches the target therefore produces no update.
While the residual is small, these are the positions where the base already agrees with the target.
Let $\mathcal{S}_t$ be the support of $\tilde p_t$, the $K$ cached tokens and $y_t$.
If the memory could fit every position exactly, the loss would approach its infimum with $s_{\theta,t}\to-\infty$ outside $\mathcal{S}_t$ and, on $\mathcal{S}_t$,
\begin{equation}
s_{\theta,t}^{\star}=\log\tilde p_t-z_{B,t}+c_t,
\label{eq:residual-opt}
\end{equation}
with a free per-position constant $c_t$.
The residual thus learns the log-ratio between the target and the base prediction.
The standalone objective of Section~\ref{sec:inference} instead drives the memory's own distribution toward $\tilde p_t$, regardless of the base.

\subsection{Inference and matched control}
\label{sec:inference}
At each decoding step, the base supplies $h_t$ and $z_{B,t}$, and the memory produces $s_{\theta,t}$.
A residual scale $\resscale\geq0$ multiplies the memory logits before the final softmax:
\begin{equation}
z_{\mathrm{final},t}=z_{B,t}+\resscale s_{\theta,t},\qquad
p_{\mathrm{final},t}=\mathrm{softmax}(z_{\mathrm{final},t}).
\label{eq:inference}
\end{equation}
This retains the additive structure used in training; $\resscale=0$ recovers the base prediction.
Evidence retrieval and teacher forwarding are offline; inference uses only the base and learned memory (cost in Appendix~\ref{app:deployment}).
At the optimum of Eq.~\ref{eq:residual-opt}, the scaled logits on $\mathcal{S}_t$ become $(1-\resscale)\,z_{B,t}+\resscale\,(\log\tilde p_t+c_t)$, and $c_t$ cancels in the softmax.
In this limit, tokens outside $\mathcal{S}_t$ receive zero probability for $\resscale>0$, and on $\mathcal{S}_t$ the scaled readout is a geometric interpolation between the base distribution $p_{B,t}=\mathrm{softmax}(z_{B,t})$ and the training target:
\begin{equation}
p_{\mathrm{final},t}\propto p_{B,t}^{\,1-\resscale}\,\tilde p_t^{\,\resscale},
\label{eq:geometric}
\end{equation}
so $\resscale$ moves the prediction along a log-linear path from Base at $\resscale=0$ to the fitted target at $\resscale=1$.

The residual can still suppress a base answer, as the condition below shows.

For the shared-teacher standalone control, we reuse the architecture, targets, examples, position masks, loss weights, and optimizer.
Its loss replaces $q_{R,t}$ with the memory-only distribution $q_{M,t}=\mathrm{softmax}(s_{\phi,t})$, where $s_{\phi,t}$ are the logits of a separately trained memory with parameters $\phi$.
Inference follows the MLP Memory probability mixture, with interpolation weight $\mixweight\in[0,1]$:
\begin{equation}
p_{\mathrm{final},t}=(1-\mixweight)\,\mathrm{softmax}(z_{B,t})
+\mixweight\,q_{M,t}.
\label{eq:mlp-readout}
\end{equation}
Both readouts recover Base at zero.
The mixture interpolates linearly between Base and the memory, and the additive readout interpolates geometrically (Eq.~\ref{eq:geometric}).
We therefore select $\mixweight$ and $\resscale$ separately for each readout (Section~\ref{sec:controlled-qa}).

\textbf{When a readout overrides the base.} Consider a position where the base prefers token $y$ to $v$, both in $\mathcal{S}_t$, with log-odds margin $m_B=\log p_{B,t}(y)-\log p_{B,t}(v)>0$, while the fitted target prefers $v$ with margin $m_T=\log\tilde p_t(v)-\log\tilde p_t(y)>0$.
By Eq.~\ref{eq:geometric}, the readout's log-odds of $v$ over $y$ are
\[
\log\frac{p_{\mathrm{final},t}(v)}{p_{\mathrm{final},t}(y)}=\resscale\, m_T-(1-\resscale)\, m_B,
\]
so the additive readout switches to $v$ if and only if
\begin{equation}
\resscale\, m_T>(1-\resscale)\, m_B,
\label{eq:override-additive}
\end{equation}
whereas the mixture of Eq.~\ref{eq:mlp-readout}, with the memory at its standalone optimum $q_{M,t}=\tilde p_t$, switches if and only if $\mixweight\,[\tilde p_t(v)-\tilde p_t(y)]>(1-\mixweight)\,[p_{B,t}(y)-p_{B,t}(v)]$.
The additive rule compares log-odds, so overriding a confident base token requires a correspondingly confident target.
The mixture compares probability gaps, and these never exceed one.
A memory that puts all its mass on $v$ therefore overrides any base margin once $\mixweight>1/2$.
Preserving confident base answers then forces a small $\mixweight$.
Under greedy decoding, this token-level condition is how a memory breaks a correct base answer (Section~\ref{sec:correction}).

\subsection{Construction storage}
\label{sec:storage-cost}
The cached targets are all that training needs from the teacher.
Both routes cache them.
The kNN route also retains a datastore and a FAISS IVF-PQ index.
The RAG route retains the retrieved text and a BM25 index instead.
Per valid position, a datastore entry holds a $d$-dimensional FP16 key and an INT32 token ID.
An index entry holds a $P$-byte PQ code and an 8-byte ID.
A cached target holds $K$ INT32 token IDs and $K$ FP16 probabilities.
Over $n$ positions, the two routes retain
\begin{equation}
\begin{aligned}
S_{\mathrm{kNN}}(n)&=(2d+4)n+(P+8)n+6Kn,\\
S_{\mathrm{RAG}}(n)&=6Kn+bn,
\end{aligned}
\label{eq:storage-scaling}
\end{equation}
where $b$ is the text and BM25 index per position.
With $d=4096$, $P=64$, and $K=64$, the kNN route needs $8196+72+384=8652$ bytes per position.
The ratio $S_{\mathrm{kNN}}/S_{\mathrm{RAG}}=8652/(384+b)$ does not depend on $n$, so the saving holds at every corpus size.
It approaches 22.5 as $b$ goes to zero, since the datastore alone takes $(2d+4)/6K\approx21.3$ times the bytes of the targets.
Section~\ref{sec:storage} measures $b$ and both routes.

\section{Experimental Setup}
\label{sec:setup}

\textbf{Models.} We use Mistral-7B-v0.3 \citep{jiang2023mistral} as the base model. 
ResMem and MLP Memory each attach an eight-layer memory with hidden size 4096, intermediate size 14336, and approximately 1.54B parameters.
Additional baselines are BM25-RAG \citep{robertson2009bm25,lewis2020rag}, LoRA \citep{hu2022lora}, and Memory Decoder \citep{cao2025memorydecoder}. 
On the Wikipedia subset, we train two memories with this architecture on a kNN teacher built from the subset.
Both start from the same initialization, follow the same example order, and train for three epochs (220.8M tokens).
One uses the standalone objective, and the other uses the residual objective (Section~\ref{sec:correction}; Appendix~\ref{app:implementation}).
For knowledge QA, the correction analysis, HaluEval, and the precision study, ResMem, MLP Memory, and Memory Decoder are each trained for one epoch on full English Wikipedia.
MLP Memory and Memory Decoder follow their official pipelines, including the kNN teacher.
The shared-teacher ablation trains eight new paired seeds for each objective and for Memory Decoder, all with equal exposure (Section~\ref{sec:controlled-qa}).

\textbf{Datasets.} Following MLP Memory, we use the December 2021 English Wikipedia dump as the corpus for memory training.
The Wikipedia subset, used for the storage comparison and these two memories, contains 74 of its 4,716 source shards, about 1/64 of the corpus.
Knowledge QA contains 38,627 questions from Natural Questions (NQ) \citep{kwiatkowski2019nq}, WebQuestions \citep{berant2013webquestions}, TriviaQA \citep{joshi2017triviaqa}, PopQA \citep{mallen2023popqa}, and HotpotQA \citep{yang2018hotpotqa}.
We also use HaluEval's Dialogue, QA, and Summarization splits \citep{li2023halueval}.
Hardware, training schedules, and evaluation configurations are given in Appendix~\ref{app:implementation}.

\section{Results}
\label{sec:results}

\begin{table}[t]
\caption{Knowledge QA on 38,627 questions. Open-domain QA averages NQ, WebQuestions, and TriviaQA; long-tail QA is PopQA; multi-hop QA is HotpotQA (per-dataset results in Appendix~\ref{app:qa-detail}). Subscripts give changes from Base: percentage points for EM and F1, and the relative reduction for PPL. Bold marks the best value per column.}
\label{tab:qa-main}
\centering\scriptsize
\setlength{\tabcolsep}{3.5pt}\renewcommand{\arraystretch}{1.12}
\resizebox{\linewidth}{!}{%
\begin{tabular}{lrrrrrrrrr}
\toprule
& \multicolumn{3}{c}{Open-domain QA} & \multicolumn{3}{c}{Long-tail QA} & \multicolumn{3}{c}{Multi-hop QA} \\
\cmidrule(lr){2-4}\cmidrule(lr){5-7}\cmidrule(lr){8-10}
Method & EM $\uparrow$ & F1 $\uparrow$ & PPL $\downarrow$ & EM $\uparrow$ & F1 $\uparrow$ & PPL $\downarrow$ & EM $\uparrow$ & F1 $\uparrow$ & PPL $\downarrow$ \\
\midrule
Base & 29.97 & 38.51 & 11.11 & 22.01 & 25.15 & 11.96 & 17.64 & 23.47 & 6.27 \\
\rowcolor{methodshade}\multicolumn{10}{c}{\textit{Non-parametric methods}} \\
+ BM25-RAG & 24.54\textsubscript{\textcolor{scoreloss}{-5.42}} & 31.87\textsubscript{\textcolor{scoreloss}{-6.64}} & 10.02\textsubscript{\textcolor{scoregain}{$\downarrow$9.82\%}} & 22.65\textsubscript{\textcolor{scoregain}{+0.64}} & 25.86\textsubscript{\textcolor{scoregain}{+0.71}} & 12.29\textsubscript{\textcolor{scoreloss}{$\uparrow$2.72\%}} & 18.51\textsubscript{\textcolor{scoregain}{+0.87}} & 26.27\textsubscript{\textcolor{scoregain}{+2.80}} & 5.88\textsubscript{\textcolor{scoregain}{$\downarrow$6.28\%}} \\
\rowcolor{methodshade}\multicolumn{10}{c}{\textit{Parametric methods}} \\
+ LoRA & 30.00\textsubscript{\textcolor{scoregain}{+0.03}} & 38.71\textsubscript{\textcolor{scoregain}{+0.20}} & 10.60\textsubscript{\textcolor{scoregain}{$\downarrow$4.67\%}} & 22.43\textsubscript{\textcolor{scoregain}{+0.42}} & 26.51\textsubscript{\textcolor{scoregain}{+1.36}} & 11.61\textsubscript{\textcolor{scoregain}{$\downarrow$2.99\%}} & 17.63\textsubscript{\textcolor{scoreloss}{-0.01}} & 25.32\textsubscript{\textcolor{scoregain}{+1.85}} & 5.93\textsubscript{\textcolor{scoregain}{$\downarrow$5.42\%}} \\
+ Memory Decoder & 30.19\textsubscript{\textcolor{scoregain}{+0.23}} & 38.95\textsubscript{\textcolor{scoregain}{+0.44}} & 10.84\textsubscript{\textcolor{scoregain}{$\downarrow$2.47\%}} & 22.12\textsubscript{\textcolor{scoregain}{+0.11}} & 25.40\textsubscript{\textcolor{scoregain}{+0.24}} & 11.80\textsubscript{\textcolor{scoregain}{$\downarrow$1.38\%}} & 17.77\textsubscript{\textcolor{scoregain}{+0.14}} & 23.76\textsubscript{\textcolor{scoregain}{+0.28}} & 6.17\textsubscript{\textcolor{scoregain}{$\downarrow$1.66\%}} \\
+ MLP Memory & 30.40\textsubscript{\textcolor{scoregain}{+0.44}} & 39.41\textsubscript{\textcolor{scoregain}{+0.90}} & 10.43\textsubscript{\textcolor{scoregain}{$\downarrow$6.12\%}} & 22.38\textsubscript{\textcolor{scoregain}{+0.37}} & 25.83\textsubscript{\textcolor{scoregain}{+0.68}} & 11.07\textsubscript{\textcolor{scoregain}{$\downarrow$7.47\%}} & 18.30\textsubscript{\textcolor{scoregain}{+0.66}} & 24.52\textsubscript{\textcolor{scoregain}{+1.05}} & 6.02\textsubscript{\textcolor{scoregain}{$\downarrow$4.01\%}} \\
\rowcolor{oursshade}+ ResMem (Ours) & \textbf{30.61}\textsubscript{\textcolor{scoregain}{+0.65}} & \textbf{39.73}\textsubscript{\textcolor{scoregain}{+1.22}} & \textbf{9.77}\textsubscript{\textcolor{scoregain}{$\downarrow$12.09\%}} & \textbf{22.85}\textsubscript{\textcolor{scoregain}{+0.84}} & \textbf{26.91}\textsubscript{\textcolor{scoregain}{+1.75}} & \textbf{10.99}\textsubscript{\textcolor{scoregain}{$\downarrow$8.13\%}} & \textbf{19.03}\textsubscript{\textcolor{scoregain}{+1.39}} & \textbf{26.29}\textsubscript{\textcolor{scoregain}{+2.82}} & \textbf{5.75}\textsubscript{\textcolor{scoregain}{$\downarrow$8.30\%}} \\
\bottomrule
\end{tabular}}
\vspace{-4mm}
\end{table}

\subsection{Knowledge QA Task Performance}
\label{sec:qa}
The QA checkpoints are each trained for one epoch on full Wikipedia (Section~\ref{sec:setup}).
We report EM, token F1, and reference-answer PPL on 38,627 questions, averaging the five datasets equally.
For each memory and dataset, we select $\mixweight$ or $\resscale$ to maximize EM on the validation set.
We then use that coefficient to compute EM, F1, and PPL on the evaluation set (Appendix~\ref{app:implementation}).
A QA-supervised tokenwise gate changes mean EM by only $+0.07$ points and mean F1 by $-0.02$ points, so ResMem is reported without it (Table~\ref{tab:gate-ablation} in Appendix~\ref{app:gate}).

ResMem has the highest EM and F1 and the lowest PPL in all three task families (Table~\ref{tab:qa-main}).
Over Base, it raises mean EM from 25.91 to 26.74 and mean F1 from 32.83 to 34.48.
Both metrics improve on every dataset (Appendix~\ref{app:qa-detail}).
Its F1 margin over MLP Memory grows from 0.32 points on open-domain QA to 1.08 on long-tail and 1.77 on multi-hop QA.
On HaluEval hallucination detection, the same checkpoints raise mean accuracy from 54.83 to 63.09, against 60.34 for MLP Memory (Appendix~\ref{app:halueval}).
Answer likelihood alone does not explain these results.
BM25-RAG lowers open-domain answer PPL below Base (10.02 versus 11.11), yet it loses 5.42 EM points.
Generation must choose its own words and decide when to stop.

\subsection{Construction storage}
\label{sec:storage}
\begin{wrapfigure}{r}{0.48\textwidth}
\centering
\vspace{-4mm}
\includegraphics[width=\linewidth]{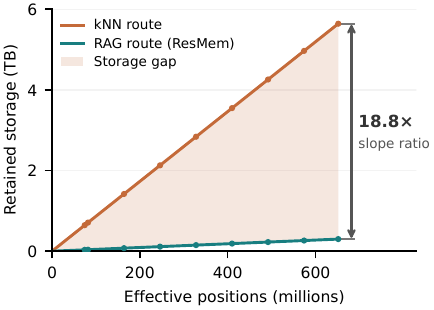}
\caption{Retained storage of the kNN and RAG routes, measured on nested shard sets of English Wikipedia up to 652M positions.}
\vspace{-4mm}
\label{fig:storage-scaling}
\end{wrapfigure}

We measure both routes of Eq.~\ref{eq:storage-scaling} on nested shard sets of the corpus, from the Wikipedia subset to 652M positions (Figure~\ref{fig:storage-scaling}).
At every set, the kNN datastore files match $(2d+4)n$ bytes.
A BM25 index and its passages, built on the same shards, grow linearly at 3.5 to 3.7 bytes per position.

The index over all of English Wikipedia takes 16.7\,GB.
On the subset, $b$ is 72.9 bytes.
The stored files add small overheads, such as span token IDs.
They take 8665 and 461 bytes per position, a ratio of 18.8.
The kNN route retains a 570.011\,GiB datastore and a 5.069\,GiB FAISS index.
The RAG route retains 5.067\,GiB of retrieved text and BM25 index.
Both cache 26.967\,GiB of targets.
The totals are 602.047 versus 32.033\,GiB, a 94.7\% ($18.8\times$) reduction in retained assets.

\subsection{Correction and harm analysis}
\label{sec:correction}
\label{sec:sparsity}
\begin{figure}[t]
\centering
\includegraphics[width=0.88\textwidth]{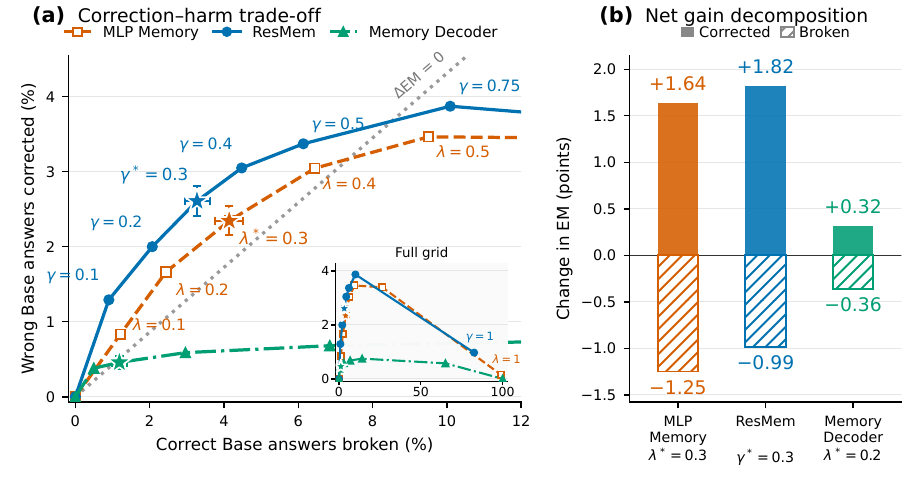}
\vspace{-8mm}
\caption{Correction--harm trade-off on 38,627 QA questions.
(a) Correction versus harm rates across coefficients; stars ($^*$) mark coefficients selected on development data, and the dotted line marks zero net gain.
(b) Correction and harm contributions at the selected coefficients.}
\label{fig:correction-harm}
\end{figure}

A change in mean EM is the net of two opposing effects: the memory fixes some questions the base model answers wrongly and breaks some it answers correctly.
We pair the answers to all 38,627 questions with the frozen Base answers and count corrections $C$ (Base wrong, memory right) and harms $H$ (Base right, memory wrong) under normalized EM.
The net EM gain is $(C-H)/Q$ over the $Q$ questions.
Each method uses one global coefficient chosen on 4,091 NQ/TriviaQA development questions disjoint from the test sets: $\mixweight^*=0.3$ for MLP Memory, $\resscale^*=0.3$ for ResMem, and $\mixweight^*=0.2$ for Memory Decoder.
Figure~\ref{fig:correction-harm}a traces correction against harm rate as the coefficient varies.
Net EM rises while a curve is steeper than the $\Delta\mathrm{EM}=0$ line.
Figure~\ref{fig:correction-harm}b splits the net gain into corrections and harms at the selected coefficients.

ResMem corrects 704 answers and harms 381, against 633 and 482 for MLP Memory.
Its net EM gain is 0.84 points, against 0.39 (paired difference 0.45, 95\% interval [0.25, 0.64]).
Most of this difference comes from the 101 avoided harms.
Memory Decoder nets $-0.04$ points, with 124 corrections and 138 harms.

\begin{figure}[t]
\centering
\includegraphics[width=0.88\linewidth]{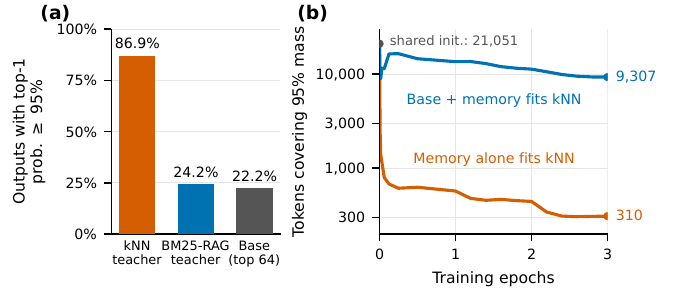}
\vspace{-4mm}
\caption{Distribution concentration.
(a) Percentage of positions whose most likely token has probability at least 95\%; the base prediction is truncated to its top 64.
(b) Tokens covering 95\% of the memory's own softmax, averaged over positions, for memories trained on the kNN teacher from a shared initialization, either alone (as in MLP Memory) or through base plus memory (as in ResMem).
The standalone memory's softmax is its mixture component.
The residual's softmax is not used at inference and is shown for reference.}
\label{fig:sparsity}
\vspace{-4mm}
\end{figure}

One source of the harm gap is the shape of the kNN target (Figure~\ref{fig:sparsity}a).
Over all 74.6M cached positions, a single token carries at least 95\% of its mass at 87\% of positions.
Its median entropy is 0, and its mean top-1 probability is 0.95.
The evidence-reading teacher is shaped like the base instead.
It does so at 24\% of 1.05M positions, and the base at 22\%.
Their mean top-1 probabilities are 0.60 and 0.59.
This sharpness is a known property of kNN-LM.
The kNN distribution comes from $k\ll|V|$ neighbors through a softmax over negative distances, so it is sharper than a model's output \citep{xu2023whyknnlm}.
A higher temperature spreads the mass \citep{khandelwal2021knnmt}.
However, it also gives distant, noisier neighbors almost the same weight as the nearest one \citep{zheng2021adaptiveknnmt}.
The original kNN-LM therefore mixes this distribution at a small weight, $\mixweight=0.25$ on WikiText-103 \citep{khandelwal2020knnlm}.
Our kNN teacher follows the MLP Memory pipeline at $T=16$.
This is flatter than the original $T=1$, so the choice favors the baselines.
Even so, the target stays sharp.

A teacher shaped like the base suits a residual.
The residual learns the log-ratio between the target and the base (Eq.~\ref{eq:residual-opt}).
Our teacher is the base itself, reading retrieved evidence.
The two share an output space and a similar shape.
The residual therefore mainly encodes what the evidence changes.
The teacher also keeps the base's uncertainty.
When several continuations are plausible, it does not force one of them.
The kNN target commits to one token at most positions.
A memory that fits it learns to commit as well.

MLP Memory shows this effect.
On the same kNN teacher, initialization, and example order, a memory trained alone narrows from 21,051 tokens covering 95\% of its mass to 310 (Figure~\ref{fig:sparsity}b).
In MLP Memory, this distribution is the component that the mixture adds to the base (Eq.~\ref{eq:mlp-readout}).
A sharp component can override a confident base answer once $\mixweight$ grows (Section~\ref{sec:inference}).
The additive readout only shifts base log-odds (Eq.~\ref{eq:override-additive}).
The residual's own softmax is not used at inference, so we judge it by the combined predictions.
Figure~\ref{fig:correction-harm} shows the expected pattern.
MLP Memory breaks 482 correct answers, and ResMem breaks 381.
The MLP Memory curve flattens earlier.
Section~\ref{sec:ablations} holds the teacher fixed to test the objective and readout alone.

\subsection{Ablations}
\label{sec:ablations}
\label{sec:controlled-qa}
\begin{table}[t]
\centering
\caption{Shared-teacher QA ablation with the BM25-RAG teacher, equal exposure, and eight paired seeds. Top: native configurations on 38,627 questions; Gold NLL is the token-weighted reference-answer NLL. Bottom: F1 at the development-selected coefficient / coefficient one on 4,096 diagnostic questions (Base 32.29).}
\label{tab:controlled-qa}
\resizebox{\linewidth}{!}{%
\begin{tabular}{lrrr}
\toprule
Native configuration & EM & F1 & Gold NLL \\
\midrule
Base & 25.91 & 32.83 & 2.295 \\
Standalone + mixture & 25.87 & 33.12 & 2.291 \\
Memory Decoder + mixture & 25.90 & 32.87 & 2.329 \\
Residual + additive & 26.03 & 33.24 & 2.273 \\
\bottomrule
\end{tabular}
\par\smallskip
\begin{tabular}{lrr}
\toprule
Training objective & Mixture readout & Additive readout \\
\midrule
Standalone & 32.65 / 21.87 & 32.29 / 28.44 \\
Residual & 32.29 / 0.06 & 32.66 / 30.68 \\
\bottomrule
\end{tabular}
}
\end{table}

The comparisons above differ in teacher, objective and readout, and memory architecture at once.
Holding the teacher fixed, we first vary the objective and readout, and then the architecture.

\textbf{Objective and readout.} Both memories use the same BM25-RAG teacher, cached inputs, eight-layer architecture, paired initialization, and example order.
Eight paired seeds each receive 8,388,608 supervised-position exposures over eight epochs.
Each configuration selects one coefficient on 1,024 development questions disjoint from the test sets.
Pairing the seeds lets each comparison share its initialization and data order.

We call the standalone objective with mixture readout and the residual objective with additive readout the native configurations.
Both improve mean F1 over Base (Table~\ref{tab:controlled-qa}).
The residual configuration leads by 0.117 points, with a 95\% question-cluster bootstrap interval of $[-0.020,0.258]$, or $[-0.180,0.477]$ when seeds are also resampled.

In the four-cell diagnostic on 4,096 shared questions, both crossed configurations select coefficient zero and recover Base.
Residual scores read as a standalone distribution yield an F1 of 0.06, so they are useful only when added to the base logits.
Pooled over the eight runs at native operating points, the residual configuration preserves 96.11\% of Base-correct answers, against 94.05\%.
It repairs 1.51\% of Base errors, against 2.25\% (point estimates).
With the teacher fixed, it thus trades some repair for preservation.

\textbf{Architecture.} Memory Decoder replaces the hidden-state memory with a separate 0.47B-parameter decoder that reads the token sequence (Appendix~\ref{app:implementation}).
With the same supervision, exposure, and optimizer schedule and a mixture readout, it reaches F1 32.87.
This is 0.26 and 0.37 points below the standalone and residual configurations ($[-0.42,-0.09]$ and $[-0.49,-0.25]$).
Development selection sets its coefficient to zero for six of the eight seeds (Table~\ref{tab:controlled-qa}).
Under this training budget and at these sizes, the hidden-state memory is the stronger design.

Unlike the shared-teacher runs, full-recipe ResMem both corrects more and harms less than MLP Memory (Section~\ref{sec:correction}).
The objective and readout thus contribute to the better preservation.
The extra repairs come with the evidence-reading teacher, in line with the sharper kNN targets in Figure~\ref{fig:sparsity}a.

\FloatBarrier

\section{Conclusion}
\label{sec:conclusion}

ResMem distills a retrieval-augmented teacher into an additive logit residual on a frozen base model.
On a Wikipedia subset, it cuts retained construction storage by 94.7\% ($18.8\times$).
On five knowledge-QA datasets, it raises mean F1 from 32.83 to 34.48.
It also corrects more base errors and harms fewer correct answers than MLP Memory.
With a shared teacher, the residual objective mainly preserves correct answers, and the extra repairs come with the evidence-reading teacher.
The storage saving and the extra repairs thus share one source, the teacher that reads retrieved text.

With only the parametric memory trained, the base model stays unchanged.
A limitation of this design is that the memory reads one hidden state and writes only to the output logits.
The teacher's gain is therefore not integrated through the base model's intermediate layers.
Memories that read from and write to several layers are a natural next step.

\subsection*{AI use statement}
In this work, generative AI tools played a supporting role, and the authors led all research decisions.
Among tasks with required disclosure, we used generative AI tools to help formalize mathematical derivations, give feedback on experiments, and help implement experiment code.
The authors developed the method and verified each formalized derivation.
The authors also reviewed and tested the AI-assisted code before using its results.
AI feedback on experiments informed decisions that the authors made.
We have not used generative AI tools to generate synthetic data, propose research ideas or hypotheses, or clean datasets.
Among tasks with recommended disclosure, we used generative AI tools to suggest the structure of the paper and to edit it for readability.
We have reviewed all AI-assisted work.
The authors checked every reported number against the stored experiment outputs and every reference against its DBLP record.
We take responsibility for the final content of this work, including text, claims, or artifacts produced with the aid of generative AI.

\subsection*{Reproducibility statement}
Section~\ref{sec:setup} and Appendix~\ref{app:implementation} describe the models, data, and evaluation settings.
Code for the ResMem training pipeline and the main evaluation pipelines is available at \plainurl{https://github.com/SSRand/From-Lookups-to-Lessons}, and the knowledge-QA checkpoint at \plainurl{https://huggingface.co/RandHan/ResMem-Mistral-7B-v0.3}.

\bibliography{iclr2026_conference}

\begin{thebibliography}{55}
\providecommand{\natexlab}[1]{#1}
\providecommand{\url}[1]{\texttt{#1}}
\expandafter\ifx\csname urlstyle\endcsname\relax
  \providecommand{\doi}[1]{doi: #1}\else
  \providecommand{\doi}{doi: \begingroup \urlstyle{rm}\Url}\fi

\bibitem[Agarwal et~al.(2024)Agarwal, Vieillard, Zhou, Stanczyk, Garea, Geist,
  and Bachem]{agarwal2024onpolicy}
Rishabh Agarwal, Nino Vieillard, Yongchao Zhou, Piotr Stanczyk, Sabela~Ramos
  Garea, Matthieu Geist, and Olivier Bachem.
\newblock On-policy distillation of language models: Learning from
  self-generated mistakes.
\newblock In \emph{{ICLR}}. OpenReview.net, 2024.

\bibitem[Behrouz et~al.(2025)Behrouz, Zhong, and Mirrokni]{behrouz2025titans}
Ali Behrouz, Peilin Zhong, and Vahab Mirrokni.
\newblock Titans: Learning to memorize at test time.
\newblock In \emph{NeurIPS}, 2025.

\bibitem[Berant et~al.(2013)Berant, Chou, Frostig, and
  Liang]{berant2013webquestions}
Jonathan Berant, Andrew Chou, Roy Frostig, and Percy Liang.
\newblock Semantic parsing on freebase from question-answer pairs.
\newblock In \emph{{EMNLP}}, pp.\  1533--1544. {ACL}, 2013.

\bibitem[Borgeaud et~al.(2022)Borgeaud, Mensch, Hoffmann, Cai, Rutherford,
  Millican, van~den Driessche, Lespiau, Damoc, Clark, de~Las~Casas, Guy,
  Menick, Ring, Hennigan, Huang, Maggiore, Jones, Cassirer, Brock, Paganini,
  Irving, Vinyals, Osindero, Simonyan, Rae, Elsen, and
  Sifre]{borgeaud2022retro}
Sebastian Borgeaud, Arthur Mensch, Jordan Hoffmann, Trevor Cai, Eliza
  Rutherford, Katie Millican, George van~den Driessche, Jean{-}Baptiste
  Lespiau, Bogdan Damoc, Aidan Clark, Diego de~Las~Casas, Aurelia Guy, Jacob
  Menick, Roman Ring, Tom Hennigan, Saffron Huang, Loren Maggiore, Chris Jones,
  Albin Cassirer, Andy Brock, Michela Paganini, Geoffrey Irving, Oriol Vinyals,
  Simon Osindero, Karen Simonyan, Jack~W. Rae, Erich Elsen, and Laurent Sifre.
\newblock Improving language models by retrieving from trillions of tokens.
\newblock In \emph{{ICML}}, volume 162 of \emph{Proceedings of Machine Learning
  Research}, pp.\  2206--2240. {PMLR}, 2022.

\bibitem[Cao et~al.(2025)Cao, Wang, Wei, Guo, Chen, Zhou, and
  Lin]{cao2025memorydecoder}
Jiaqi Cao, Jiarui Wang, Rubin Wei, Qipeng Guo, Kai Chen, Bowen Zhou, and
  Zhouhan Lin.
\newblock Memory decoder: {A} pretrained, plug-and-play memory for large
  language models.
\newblock In \emph{NeurIPS}, 2025.

\bibitem[Cheng et~al.(2024)Cheng, Wang, Zhang, Ge, Chen, Wei, Zhang, and
  Zhao]{cheng2024xrag}
Xin Cheng, Xun Wang, Xingxing Zhang, Tao Ge, Si{-}Qing Chen, Furu Wei, Huishuai
  Zhang, and Dongyan Zhao.
\newblock xrag: Extreme context compression for retrieval-augmented generation
  with one token.
\newblock In \emph{NeurIPS}, 2024.

\bibitem[Eyuboglu et~al.(2025)Eyuboglu, Ehrlich, Arora, Guha, Zinsley, Liu,
  Tennien, Rudra, Zou, Mirhoseini, and R{\'{e}}]{eyuboglu2025cartridges}
Sabri Eyuboglu, Ryan Ehrlich, Simran Arora, Neel Guha, Dylan Zinsley, Emily
  Liu, Will Tennien, Atri Rudra, James Zou, Azalia Mirhoseini, and Christopher
  R{\'{e}}.
\newblock Cartridges: Lightweight and general-purpose long context
  representations via self-study.
\newblock \emph{CoRR}, abs/2506.06266, 2025.

\bibitem[Furlanello et~al.(2018)Furlanello, Lipton, Tschannen, Itti, and
  Anandkumar]{furlanello2018bornagain}
Tommaso Furlanello, Zachary~Chase Lipton, Michael Tschannen, Laurent Itti, and
  Anima Anandkumar.
\newblock Born-again neural networks.
\newblock In \emph{{ICML}}, volume~80 of \emph{Proceedings of Machine Learning
  Research}, pp.\  1602--1611. {PMLR}, 2018.

\bibitem[Geva et~al.(2021)Geva, Schuster, Berant, and Levy]{geva2021ffn}
Mor Geva, Roei Schuster, Jonathan Berant, and Omer Levy.
\newblock Transformer feed-forward layers are key-value memories.
\newblock In \emph{{EMNLP} {(1)}}, pp.\  5484--5495. Association for
  Computational Linguistics, 2021.

\bibitem[He et~al.(2021)He, Neubig, and Berg{-}Kirkpatrick]{he2021efficientknn}
Junxian He, Graham Neubig, and Taylor Berg{-}Kirkpatrick.
\newblock Efficient nearest neighbor language models.
\newblock In \emph{{EMNLP} {(1)}}, pp.\  5703--5714. Association for
  Computational Linguistics, 2021.

\bibitem[Hinton et~al.(2015)Hinton, Vinyals, and Dean]{hinton2015distilling}
Geoffrey~E. Hinton, Oriol Vinyals, and Jeffrey Dean.
\newblock Distilling the knowledge in a neural network.
\newblock \emph{CoRR}, abs/1503.02531, 2015.

\bibitem[Hu et~al.(2022)Hu, Shen, Wallis, Allen{-}Zhu, Li, Wang, Wang, and
  Chen]{hu2022lora}
Edward~J. Hu, Yelong Shen, Phillip Wallis, Zeyuan Allen{-}Zhu, Yuanzhi Li,
  Shean Wang, Lu~Wang, and Weizhu Chen.
\newblock Lora: Low-rank adaptation of large language models.
\newblock In \emph{{ICLR}}. OpenReview.net, 2022.

\bibitem[Izacard et~al.(2023)Izacard, Lewis, Lomeli, Hosseini, Petroni, Schick,
  Dwivedi{-}Yu, Joulin, Riedel, and Grave]{izacard2022atlas}
Gautier Izacard, Patrick Lewis, Maria Lomeli, Lucas Hosseini, Fabio Petroni,
  Timo Schick, Jane Dwivedi{-}Yu, Armand Joulin, Sebastian Riedel, and Edouard
  Grave.
\newblock Atlas: Few-shot learning with retrieval augmented language models.
\newblock \emph{J. Mach. Learn. Res.}, 24:\penalty0 251:1--251:43, 2023.

\bibitem[J{\'{e}}gou et~al.(2011)J{\'{e}}gou, Douze, and Schmid]{jegou2011pq}
Herv{\'{e}} J{\'{e}}gou, Matthijs Douze, and Cordelia Schmid.
\newblock Product quantization for nearest neighbor search.
\newblock \emph{{IEEE} Trans. Pattern Anal. Mach. Intell.}, 33\penalty0
  (1):\penalty0 117--128, 2011.

\bibitem[Jiang et~al.(2023)Jiang, Sablayrolles, Mensch, Bamford, Chaplot,
  de~Las~Casas, Bressand, Lengyel, Lample, Saulnier, Lavaud, Lachaux, Stock,
  Scao, Lavril, Wang, Lacroix, and Sayed]{jiang2023mistral}
Albert~Q. Jiang, Alexandre Sablayrolles, Arthur Mensch, Chris Bamford,
  Devendra~Singh Chaplot, Diego de~Las~Casas, Florian Bressand, Gianna Lengyel,
  Guillaume Lample, Lucile Saulnier, L{\'{e}}lio~Renard Lavaud, Marie{-}Anne
  Lachaux, Pierre Stock, Teven~Le Scao, Thibaut Lavril, Thomas Wang,
  Timoth{\'{e}}e Lacroix, and William~El Sayed.
\newblock Mistral 7b.
\newblock \emph{CoRR}, abs/2310.06825, 2023.

\bibitem[Johnson et~al.(2021)Johnson, Douze, and J{\'{e}}gou]{johnson2021faiss}
Jeff Johnson, Matthijs Douze, and Herv{\'{e}} J{\'{e}}gou.
\newblock Billion-scale similarity search with gpus.
\newblock \emph{{IEEE} Trans. Big Data}, 7\penalty0 (3):\penalty0 535--547,
  2021.

\bibitem[Joshi et~al.(2017)Joshi, Choi, Weld, and
  Zettlemoyer]{joshi2017triviaqa}
Mandar Joshi, Eunsol Choi, Daniel~S. Weld, and Luke Zettlemoyer.
\newblock Triviaqa: {A} large scale distantly supervised challenge dataset for
  reading comprehension.
\newblock In \emph{{ACL} {(1)}}, pp.\  1601--1611. Association for
  Computational Linguistics, 2017.

\bibitem[Khandelwal et~al.(2020)Khandelwal, Levy, Jurafsky, Zettlemoyer, and
  Lewis]{khandelwal2020knnlm}
Urvashi Khandelwal, Omer Levy, Dan Jurafsky, Luke Zettlemoyer, and Mike Lewis.
\newblock Generalization through memorization: Nearest neighbor language
  models.
\newblock In \emph{{ICLR}}. OpenReview.net, 2020.

\bibitem[Khandelwal et~al.(2021)Khandelwal, Fan, Jurafsky, Zettlemoyer, and
  Lewis]{khandelwal2021knnmt}
Urvashi Khandelwal, Angela Fan, Dan Jurafsky, Luke Zettlemoyer, and Mike Lewis.
\newblock Nearest neighbor machine translation.
\newblock In \emph{{ICLR}}. OpenReview.net, 2021.

\bibitem[Kwiatkowski et~al.(2019)Kwiatkowski, Palomaki, Redfield, Collins,
  Parikh, Alberti, Epstein, Polosukhin, Devlin, Lee, Toutanova, Jones, Kelcey,
  Chang, Dai, Uszkoreit, Le, and Petrov]{kwiatkowski2019nq}
Tom Kwiatkowski, Jennimaria Palomaki, Olivia Redfield, Michael Collins,
  Ankur~P. Parikh, Chris Alberti, Danielle Epstein, Illia Polosukhin, Jacob
  Devlin, Kenton Lee, Kristina Toutanova, Llion Jones, Matthew Kelcey,
  Ming{-}Wei Chang, Andrew~M. Dai, Jakob Uszkoreit, Quoc Le, and Slav Petrov.
\newblock Natural questions: a benchmark for question answering research.
\newblock \emph{Trans. Assoc. Comput. Linguistics}, 7:\penalty0 452--466, 2019.

\bibitem[Lei et~al.(2026{\natexlab{a}})Lei, Liu, Chen, Cai, Shi, Wang, and
  Wang]{lei2026transmem}
Haodong Lei, Junming Liu, Yirong Chen, Pinlong Cai, Botian Shi, Ding Wang, and
  Hongsong Wang.
\newblock Transmem: Transforming hidden states into memory for large language
  models.
\newblock \emph{CoRR}, abs/2607.29032, 2026{\natexlab{a}}.

\bibitem[Lei et~al.(2026{\natexlab{b}})Lei, Zhang, Li, Wang, Fan, Liu, Liu, Ma,
  Chen, and Poria]{lei2026deltamem}
Jingdi Lei, Di~Zhang, Junxian Li, Weida Wang, Kaixuan Fan, Xiang Liu, Qihan
  Liu, Xiaoteng Ma, Baian Chen, and Soujanya Poria.
\newblock {\(\delta\)}-mem: Efficient online memory for large language models.
\newblock \emph{CoRR}, abs/2605.12357, 2026{\natexlab{b}}.

\bibitem[Lewis et~al.(2020)Lewis, Perez, Piktus, Petroni, Karpukhin, Goyal,
  K{\"{u}}ttler, Lewis, Yih, Rockt{\"{a}}schel, Riedel, and
  Kiela]{lewis2020rag}
Patrick Lewis, Ethan Perez, Aleksandra Piktus, Fabio Petroni, Vladimir
  Karpukhin, Naman Goyal, Heinrich K{\"{u}}ttler, Mike Lewis, Wen{-}tau Yih,
  Tim Rockt{\"{a}}schel, Sebastian Riedel, and Douwe Kiela.
\newblock Retrieval-augmented generation for knowledge-intensive {NLP} tasks.
\newblock In \emph{NeurIPS}, 2020.

\bibitem[Li et~al.(2023{\natexlab{a}})Li, Cheng, Zhao, Nie, and
  Wen]{li2023halueval}
Junyi Li, Xiaoxue Cheng, Xin Zhao, Jian{-}Yun Nie, and Ji{-}Rong Wen.
\newblock Halueval: {A} large-scale hallucination evaluation benchmark for
  large language models.
\newblock In \emph{{EMNLP}}, pp.\  6449--6464. Association for Computational
  Linguistics, 2023{\natexlab{a}}.

\bibitem[Li et~al.(2023{\natexlab{b}})Li, Holtzman, Fried, Liang, Eisner,
  Hashimoto, Zettlemoyer, and Lewis]{li2023contrastive}
Xiang~Lisa Li, Ari Holtzman, Daniel Fried, Percy Liang, Jason Eisner, Tatsunori
  Hashimoto, Luke Zettlemoyer, and Mike Lewis.
\newblock Contrastive decoding: Open-ended text generation as optimization.
\newblock In \emph{{ACL} {(1)}}, pp.\  12286--12312. Association for
  Computational Linguistics, 2023{\natexlab{b}}.

\bibitem[Liu et~al.(2021)Liu, Sap, Lu, Swayamdipta, Bhagavatula, Smith, and
  Choi]{liu2021dexperts}
Alisa Liu, Maarten Sap, Ximing Lu, Swabha Swayamdipta, Chandra Bhagavatula,
  Noah~A. Smith, and Yejin Choi.
\newblock Dexperts: Decoding-time controlled text generation with experts and
  anti-experts.
\newblock In \emph{{ACL/IJCNLP} {(1)}}, pp.\  6691--6706. Association for
  Computational Linguistics, 2021.

\bibitem[Liu et~al.(2024{\natexlab{a}})Liu, Han, Wang, Tsvetkov, Choi, and
  Smith]{liu2024proxy}
Alisa Liu, Xiaochuang Han, Yizhong Wang, Yulia Tsvetkov, Yejin Choi, and
  Noah~A. Smith.
\newblock Tuning language models by proxy.
\newblock \emph{CoRR}, abs/2401.08565, 2024{\natexlab{a}}.

\bibitem[Liu et~al.(2024{\natexlab{b}})Liu, Lin, Hewitt, Paranjape, Bevilacqua,
  Petroni, and Liang]{liu2024lostmiddle}
Nelson~F. Liu, Kevin Lin, John Hewitt, Ashwin Paranjape, Michele Bevilacqua,
  Fabio Petroni, and Percy Liang.
\newblock Lost in the middle: How language models use long contexts.
\newblock \emph{Trans. Assoc. Comput. Linguistics}, 12:\penalty0 157--173,
  2024{\natexlab{b}}.

\bibitem[Lopez{-}Paz et~al.(2016)Lopez{-}Paz, Bottou, Sch{\"{o}}lkopf, and
  Vapnik]{lopezpaz2016privileged}
David Lopez{-}Paz, L{\'{e}}on Bottou, Bernhard Sch{\"{o}}lkopf, and Vladimir
  Vapnik.
\newblock Unifying distillation and privileged information.
\newblock In \emph{{ICLR} (Poster)}, 2016.

\bibitem[Lyu et~al.(2026)Lyu, Zhong, Chen, Ruan, Liu, Lv, Wen, Wong, and
  Liang]{lyu2026tsmemory}
Sisuo Lyu, Siru Zhong, Tiegang Chen, Weilin Ruan, Qingxiang Liu, Taiqiang Lv,
  Qingsong Wen, Raymond~Chi{-}Wing Wong, and Yuxuan Liang.
\newblock Ts-memory: Plug-and-play memory for time series foundation models.
\newblock \emph{CoRR}, abs/2602.11550, 2026.

\bibitem[Mallen et~al.(2023)Mallen, Asai, Zhong, Das, Khashabi, and
  Hajishirzi]{mallen2023popqa}
Alex Mallen, Akari Asai, Victor Zhong, Rajarshi Das, Daniel Khashabi, and
  Hannaneh Hajishirzi.
\newblock When not to trust language models: Investigating effectiveness of
  parametric and non-parametric memories.
\newblock In \emph{{ACL} {(1)}}, pp.\  9802--9822. Association for
  Computational Linguistics, 2023.

\bibitem[Packer et~al.(2023)Packer, Fang, Patil, Lin, Wooders, and
  Gonzalez]{packer2024memgptllmsoperatingsystems}
Charles Packer, Vivian Fang, Shishir~G. Patil, Kevin Lin, Sarah Wooders, and
  Joseph~E. Gonzalez.
\newblock Memgpt: Towards llms as operating systems.
\newblock \emph{CoRR}, abs/2310.08560, 2023.

\bibitem[Park et~al.(2023)Park, O'Brien, Cai, Morris, Liang, and
  Bernstein]{park2023generativeagents}
Joon~Sung Park, Joseph~C. O'Brien, Carrie~Jun Cai, Meredith~Ringel Morris,
  Percy Liang, and Michael~S. Bernstein.
\newblock Generative agents: Interactive simulacra of human behavior.
\newblock In \emph{{UIST}}, pp.\  2:1--2:22. {ACM}, 2023.

\bibitem[Robertson \& Zaragoza(2009)Robertson and Zaragoza]{robertson2009bm25}
Stephen~E. Robertson and Hugo Zaragoza.
\newblock The probabilistic relevance framework: {BM25} and beyond.
\newblock \emph{Found. Trends Inf. Retr.}, 3\penalty0 (4):\penalty0 333--389,
  2009.

\bibitem[Shi et~al.(2024)Shi, Han, Lewis, Tsvetkov, Zettlemoyer, and
  Yih]{shi2024cad}
Weijia Shi, Xiaochuang Han, Mike Lewis, Yulia Tsvetkov, Luke Zettlemoyer, and
  Wen{-}tau Yih.
\newblock Trusting your evidence: Hallucinate less with context-aware decoding.
\newblock In \emph{{NAACL} (Short Papers)}, pp.\  783--791. Association for
  Computational Linguistics, 2024.

\bibitem[Snell et~al.(2022)Snell, Klein, and Zhong]{snell2022context}
Charlie Snell, Dan Klein, and Ruiqi Zhong.
\newblock Learning by distilling context.
\newblock \emph{CoRR}, abs/2209.15189, 2022.

\bibitem[Srinivasan et~al.(2022)Srinivasan, Raman, Samanta, Liao, Bertelli, and
  Bendersky]{srinivasan2022quill}
Krishna Srinivasan, Karthik Raman, Anupam Samanta, Lingrui Liao, Luca Bertelli,
  and Michael Bendersky.
\newblock {QUILL:} query intent with large language models using retrieval
  augmentation and multi-stage distillation.
\newblock In \emph{{EMNLP} (Industry Track)}, pp.\  492--501. Association for
  Computational Linguistics, 2022.

\bibitem[Su et~al.(2025)Su, Tang, Ai, Yan, Wang, Wang, Ye, Zhou, and
  Liu]{su2025parametricrag}
Weihang Su, Yichen Tang, Qingyao Ai, Junxi Yan, Changyue Wang, Hongning Wang,
  Ziyi Ye, Yujia Zhou, and Yiqun Liu.
\newblock Parametric retrieval augmented generation.
\newblock In \emph{{SIGIR}}, pp.\  1240--1250. {ACM}, 2025.

\bibitem[Sun et~al.(2025)Sun, Li, Dalal, Xu, Vikram, Zhang, Dubois, Chen, Wang,
  Koyejo, Hashimoto, and Guestrin]{sun2025ttt}
Yu~Sun, Xinhao Li, Karan Dalal, Jiarui Xu, Arjun Vikram, Genghan Zhang, Yann
  Dubois, Xinlei Chen, Xiaolong Wang, Sanmi Koyejo, Tatsunori Hashimoto, and
  Carlos Guestrin.
\newblock Learning to (learn at test time): Rnns with expressive hidden states.
\newblock In \emph{{ICML}}, volume 267 of \emph{Proceedings of Machine Learning
  Research}. {PMLR} / OpenReview.net, 2025.

\bibitem[Wang et~al.(2024)Wang, Ping, McAfee, Xu, Li, Shoeybi, and
  Catanzaro]{wang2024instructretro}
Boxin Wang, Wei Ping, Lawrence McAfee, Peng Xu, Bo~Li, Mohammad Shoeybi, and
  Bryan Catanzaro.
\newblock Instructretro: Instruction tuning post retrieval-augmented
  pretraining.
\newblock In \emph{{ICML}}, volume 235 of \emph{Proceedings of Machine Learning
  Research}, pp.\  51255--51272. {PMLR} / OpenReview.net, 2024.

\bibitem[Wang et~al.(2026{\natexlab{a}})Wang, Shi, Cao, Wei, Wang, Sun, Wang,
  Yang, Guo, Zhou, and Lin]{wang2026memsft}
Jiarui Wang, Xiang Shi, Jiaqi Cao, Rubin Wei, Xiquan Wang, Hao Sun, Jingzhi
  Wang, Zhiqi Yang, Qipeng Guo, Bowen Zhou, and Zhouhan Lin.
\newblock Memsft: Mitigating alignment tax with an external parametric memory.
\newblock \emph{CoRR}, abs/2607.25614, 2026{\natexlab{a}}.

\bibitem[Wang et~al.(2023)Wang, Dong, Cheng, Liu, Yan, Gao, and
  Wei]{wang2023longmem}
Weizhi Wang, Li~Dong, Hao Cheng, Xiaodong Liu, Xifeng Yan, Jianfeng Gao, and
  Furu Wei.
\newblock Augmenting language models with long-term memory.
\newblock In \emph{NeurIPS}, 2023.

\bibitem[Wang et~al.(2026{\natexlab{b}})Wang, Dang, Zhu, Wen, Fu, Chai, and
  Lee]{wang2026ttcd}
Zixuan Wang, Xingyu Dang, Rui{-}Jie Zhu, Zixin Wen, Hengyu Fu, Wenhao Chai, and
  Jason~D. Lee.
\newblock Learning what to remember: Test-time training via context
  distillation.
\newblock \emph{CoRR}, abs/2608.01672, 2026{\natexlab{b}}.

\bibitem[Wei et~al.(2025)Wei, Cao, Wang, Kai, Guo, Zhou, and
  Lin]{wei2025mlpmemory}
Rubin Wei, Jiaqi Cao, Jiarui Wang, Jushi Kai, Qipeng Guo, Bowen Zhou, and
  Zhouhan Lin.
\newblock {MLP} memory: {A} retriever-pretrained memory for large language
  models.
\newblock \emph{CoRR}, abs/2508.01832, 2025.

\bibitem[Wei et~al.(2026)Wei, Cao, Wang, Zhang, Guo, Zhou, and
  Lin]{wei2026memorydecoderscale}
Rubin Wei, Jiaqi Cao, Jiarui Wang, Junming Zhang, Qipeng Guo, Bowen Zhou, and
  Zhouhan Lin.
\newblock Memory decoder at scale: {A} pretrained, parametric long-term memory.
\newblock \emph{CoRR}, abs/2607.27919, 2026.

\bibitem[Wu et~al.(2022)Wu, Rabe, Hutchins, and Szegedy]{wu2022memorizing}
Yuhuai Wu, Markus~Norman Rabe, DeLesley Hutchins, and Christian Szegedy.
\newblock Memorizing transformers.
\newblock In \emph{{ICLR}}. OpenReview.net, 2022.

\bibitem[Xu et~al.(2023)Xu, Alon, and Neubig]{xu2023whyknnlm}
Frank~F. Xu, Uri Alon, and Graham Neubig.
\newblock Why do nearest neighbor language models work?
\newblock In \emph{{ICML}}, volume 202 of \emph{Proceedings of Machine Learning
  Research}, pp.\  38325--38341. {PMLR}, 2023.

\bibitem[Yang et~al.(2018)Yang, Qi, Zhang, Bengio, Cohen, Salakhutdinov, and
  Manning]{yang2018hotpotqa}
Zhilin Yang, Peng Qi, Saizheng Zhang, Yoshua Bengio, William~W. Cohen, Ruslan
  Salakhutdinov, and Christopher~D. Manning.
\newblock Hotpotqa: {A} dataset for diverse, explainable multi-hop question
  answering.
\newblock In \emph{{EMNLP}}, pp.\  2369--2380. Association for Computational
  Linguistics, 2018.

\bibitem[Yang et~al.(2022)Yang, Sun, and Wan]{yang2022knnkd}
Zhixian Yang, Renliang Sun, and Xiaojun Wan.
\newblock Nearest neighbor knowledge distillation for neural machine
  translation.
\newblock In \emph{{NAACL-HLT}}, pp.\  5546--5556. Association for
  Computational Linguistics, 2022.

\bibitem[Zhang \& Sennrich(2019)Zhang and Sennrich]{zhang2019rmsnorm}
Biao Zhang and Rico Sennrich.
\newblock Root mean square layer normalization.
\newblock In \emph{NeurIPS}, pp.\  12360--12371, 2019.

\bibitem[Zhang et~al.(2026)Zhang, Ding, Pan, Yang, Kang, Xiong, and
  Gu]{zhang2026opsdl}
Xinsen Zhang, Zhenkai Ding, Tianjun Pan, Run Yang, Chun Kang, Xue Xiong, and
  Jingnan Gu.
\newblock {OPSDL:} on-policy self-distillation for long-context language
  models.
\newblock \emph{CoRR}, abs/2604.17535, 2026.

\bibitem[Zhang et~al.(2025)Zhang, Dai, Bo, Ma, Li, Chen, Zhu, Dong, and
  Wen]{zhang2025memorysurvey}
Zeyu Zhang, Quanyu Dai, Xiaohe Bo, Chen Ma, Rui Li, Xu~Chen, Jieming Zhu,
  Zhenhua Dong, and Ji{-}Rong Wen.
\newblock A survey on the memory mechanism of large language model-based
  agents.
\newblock \emph{{ACM} Trans. Inf. Syst.}, 43\penalty0 (6):\penalty0
  155:1--155:47, 2025.

\bibitem[Zhao et~al.(2026)Zhao, Xie, Liu, Huang, Pang, Chen, and
  Grover]{zhao2026opsd}
Siyan Zhao, Zhihui Xie, Mengchen Liu, Jing Huang, Guan Pang, Feiyu Chen, and
  Aditya Grover.
\newblock Self-distilled reasoner: On-policy self-distillation for large
  language models.
\newblock \emph{CoRR}, abs/2601.18734, 2026.

\bibitem[Zheng et~al.(2021)Zheng, Zhang, Guo, Huang, Chen, Luo, and
  Chen]{zheng2021adaptiveknnmt}
Xin Zheng, Zhirui Zhang, Junliang Guo, Shujian Huang, Boxing Chen, Weihua Luo,
  and Jiajun Chen.
\newblock Adaptive nearest neighbor machine translation.
\newblock In \emph{{ACL/IJCNLP} {(2)}}, pp.\  368--374. Association for
  Computational Linguistics, 2021.

\bibitem[Zhu et~al.(2024)Zhu, Liao, Xu, Guan, Wang, and Wang]{zhu2024ra2fd}
Zhiyuan Zhu, Yusheng Liao, Chenxin Xu, Yunfeng Guan, Yanfeng Wang, and Yu~Wang.
\newblock {RA2FD:} distilling faithfulness into efficient dialogue systems.
\newblock In \emph{{EMNLP}}, pp.\  12304--12317. Association for Computational
  Linguistics, 2024.

\end{thebibliography}
\bibliographystyle{iclr2027_conference}

\clearpage
\appendix
\makeatletter\setlength{\@fptop}{0pt}\makeatother
\section{Implementation details}
\label{app:implementation}
\textbf{Memory input.} In the Mistral backbone, $h_t$ is the state after the attention residual connection and its following RMS normalization \citep{zhang2019rmsnorm}.
Base logits are computed separately through the remaining feed-forward computation, final normalization, and frozen vocabulary head.

\textbf{Subset kNN teacher and Memory Decoder.} The kNN teacher for the Wikipedia subset is built with the original teacher's builder.
It reproduces the original position count and 26.967\,GiB target cache.
Memory Decoder uses a Qwen3-0.6B-Base body with the Mistral vocabulary (0.47B parameters).

\textbf{Knowledge-QA configuration.} For each dataset, we select $\mixweight$ for MLP Memory and Memory Decoder and $\resscale$ for ResMem by the highest EM on the validation set.
Each memory then uses its selected coefficient for all three metrics on the evaluation set: EM, F1, and PPL, with EM and F1 computed from the same generated answers.
LoRA uses strength 0.25, chosen on development data.
BM25-RAG retrieves five Wikipedia passages with a 4096-token input and up to 64 output tokens; the other methods use 384 input tokens and at most 12 output tokens.
PPL scores answer tokens and excludes the prompt and EOS.
On a separate Wikipedia sample, with coefficients selected on 128 calibration sequences and scored on 256 others, the QA checkpoints reach perplexity 7.441 (MLP Memory), 8.660 (ResMem), and 9.959 (Base).
MLP Memory fits held-out text more closely, yet ResMem answers more questions correctly (Table~\ref{tab:qa-main}).
A closer fit to the corpus thus does not by itself give better final answers.

\textbf{Hardware.} The two memories trained on the subset's kNN teacher use A100-SXM4-80GB GPUs.
Knowledge-QA memory sweeps use NVIDIA A100-SXM4-40GB GPUs.
QA inference uses BF16 base computation, eager attention, batch size 16, and greedy decoding.
The shared-teacher ablation and precision measurements use two eight-GPU A100-SXM4-80GB hosts, with one model worker per GPU.
The precision study uses BF16 Base with FP32 or BF16 memory and SDPA/KV caching.
Deployment timing uses one GPU.
Detailed schedules, checkpoint sources, and scoring configurations are recorded in the repository's experimental-protocol note.

\section{Per-dataset knowledge-QA results}
\label{app:qa-detail}
Table~\ref{tab:qa-detail} lists the values behind Table~\ref{tab:qa-main}. ResMem improves EM and F1 over Base on every dataset and has the lowest PPL on each. MLP Memory is slightly higher on WebQuestions EM and F1, the smallest dataset.

\begin{table}[!htbp]
\caption{Per-dataset knowledge-QA results behind Table~\ref{tab:qa-main}, whose group values are means of these columns before rounding. Coefficients are selected per dataset (Appendix~\ref{app:implementation}). Bold marks the best value per column.}
\label{tab:qa-detail}
\centering
\footnotesize
\setlength{\tabcolsep}{3.5pt}
\begin{tabular}{lrrrrr}
\toprule
& NQ & WebQ & TriviaQA & PopQA & HotpotQA \\
Method & ($n$=3,610) & ($n$=2,032) & ($n$=11,313) & ($n$=14,267) & ($n$=7,405) \\
\midrule
\rowcolor{methodshade}\multicolumn{6}{c}{\textit{EM $\uparrow$}} \\
Base & 18.31 & 16.88 & 54.71 & 22.01 & 17.64 \\
+ BM25-RAG & 15.87 & 7.48 & 50.28 & 22.65 & 18.51 \\
+ LoRA & 18.55 & 16.73 & 54.71 & 22.43 & 17.63 \\
+ Memory Decoder & 18.50 & 17.18 & 54.90 & 22.12 & 17.77 \\
+ MLP Memory & 18.59 & \textbf{17.37} & 55.26 & 22.38 & 18.30 \\
\rowcolor{oursshade}+ ResMem (Ours) & \textbf{18.98} & 17.32 & \textbf{55.55} & \textbf{22.85} & \textbf{19.03} \\
\rowcolor{methodshade}\multicolumn{6}{c}{\textit{F1 $\uparrow$}} \\
Base & 24.52 & 30.77 & 60.24 & 25.15 & 23.47 \\
+ BM25-RAG & 21.89 & 16.76 & 56.95 & 25.86 & 26.27 \\
+ LoRA & 24.56 & 30.92 & 60.64 & 26.51 & 25.32 \\
+ Memory Decoder & 24.88 & 31.47 & 60.49 & 25.40 & 23.76 \\
+ MLP Memory & 25.11 & \textbf{32.08} & 61.05 & 25.83 & 24.52 \\
\rowcolor{oursshade}+ ResMem (Ours) & \textbf{25.69} & 31.94 & \textbf{61.57} & \textbf{26.91} & \textbf{26.29} \\
\rowcolor{methodshade}\multicolumn{6}{c}{\textit{PPL $\downarrow$}} \\
Base & 8.46 & 10.57 & 14.31 & 11.96 & 6.27 \\
+ BM25-RAG & 7.88 & 9.14 & 13.05 & 12.29 & 5.88 \\
+ LoRA & 8.08 & 10.17 & 13.54 & 11.61 & 5.93 \\
+ Memory Decoder & 8.32 & 10.25 & 13.95 & 11.80 & 6.17 \\
+ MLP Memory & 8.07 & 9.54 & 13.69 & 11.07 & 6.02 \\
\rowcolor{oursshade}+ ResMem (Ours) & \textbf{7.62} & \textbf{8.81} & \textbf{12.89} & \textbf{10.99} & \textbf{5.75} \\
\bottomrule
\end{tabular}
\end{table}

\section{Hallucination detection}
\label{app:halueval}
\label{sec:halueval}
Knowledge stored in the memory may also help the model judge whether a given answer is supported.
HaluEval tests this with few-shot Yes/No hallucination detection.
We reuse the QA checkpoints and method-specific coefficients without task-specific tuning; BM25-RAG supplies five Wikipedia passages.
ResMem raises mean detection accuracy from 54.83 for Base to 63.09, compared with 60.34 for MLP Memory (Table~\ref{tab:halueval}).
Its margin over MLP Memory is 7.00 points on QA and 1.48 on summarization, while MLP Memory is 0.22 points higher on dialogue.

\begin{table}[!htbp]
\caption{HaluEval hallucination-detection accuracy (\%). Average weights the three tasks equally. Subscripts denote percentage-point changes from the base model; bold marks the column best.}
\label{tab:halueval}
\centering\scriptsize
\setlength{\tabcolsep}{3.5pt}\renewcommand{\arraystretch}{1.12}
\resizebox{.90\linewidth}{!}{%
\begin{tabular}{lrrrr}
\toprule
Method & Dialogue & QA & Summarization & Average \\
\midrule
Mistral-7B-v0.3 & 58.02 & 56.04 & 50.42 & 54.83 \\
+ BM25-RAG & 63.57\textsubscript{\textcolor{scoregain}{+5.55}} & 52.75\textsubscript{\textcolor{scoreloss}{-3.29}} & 50.74\textsubscript{\textcolor{scoregain}{+0.32}} & 55.69\textsubscript{\textcolor{scoregain}{+0.86}} \\
+ MLP Memory & \textbf{65.48}\textsubscript{\textcolor{scoregain}{+7.46}} & 63.08\textsubscript{\textcolor{scoregain}{+7.04}} & 52.45\textsubscript{\textcolor{scoregain}{+2.03}} & 60.34\textsubscript{\textcolor{scoregain}{+5.51}} \\
\rowcolor{oursshade}+ ResMem (Ours) & 65.26\textsubscript{\textcolor{scoregain}{+7.24}} & \textbf{70.08}\textsubscript{\textcolor{scoregain}{+14.04}} & \textbf{53.93}\textsubscript{\textcolor{scoregain}{+3.51}} & \textbf{63.09}\textsubscript{\textcolor{scoregain}{+8.26}} \\
\bottomrule
\end{tabular}}
\end{table}

\section{Precision and deployment cost}
\label{app:deployment}
The construction saving leaves open how much computation the memory adds at inference and whether lower precision can reduce it.
We load the knowledge-QA checkpoints with FP32 or BF16 memory, keeping a BF16 Base and fixed coefficients $\mixweight=0.25$ and $\resscale=0.4$ for all datasets.
Across all 38,627 questions, BF16 changes mean F1 by $-0.0159$ points for ResMem (95\% paired question-cluster interval $[-0.0539,0.0195]$), $+0.0227$ for MLP Memory ($[-0.0195,0.0684]$), and $+0.0129$ for Memory Decoder ($[-0.0063,0.0338]$).

BF16 loading lowers latency and memory (Table~\ref{tab:memory-deployment}).
At batch 16, BF16 ResMem reduces fixed-length total latency from 1,254 to 1,098\,ms and raises decode throughput from 595 to 727 tokens/s.
Its additional peak allocated memory over Base falls from 5.750 to 2.875\,GiB.
It adds 8.6\% latency over Base, or 16.3\% at batch one.
Memory Decoder is smaller (0.47B parameters) and adds only 1.01\,GiB of peak memory in BF16, but it runs its own 28-layer Transformer with a KV cache at every step, adding 63.5\% latency over Base measured on the same host.
The timing experiment fixes 384 input and 32 output tokens with SDPA and KV caching.

\begin{table}[!htbp]
\centering
\small
\setlength{\tabcolsep}{5pt}
\caption{Precision and deployment cost for the knowledge-QA checkpoints at fixed $\mixweight=0.25$ (MLP Memory and Memory Decoder) and $\resscale=0.4$ (ResMem) on every dataset; because Table~\ref{tab:qa-main} selects coefficients per dataset by EM and decodes with eager attention, its F1 values differ. Quality covers 38,627 questions with EOS-or-12-token decoding; timing uses batch 16, 384 input tokens, and exactly 32 output tokens on one A100-SXM4-80GB with BF16 Base, SDPA and KV caching. Times and peak allocated memory are medians over ten steady batches; decode throughput uses pooled generated tokens per second. Memory Decoder rows were measured on a second identical host, where Base takes 1,014\,ms in total. These checkpoints and runtime differ from Table~\ref{tab:controlled-qa}.}
\label{tab:memory-deployment}
\begin{tabular}{llrrrrr}
\toprule
Method & Memory & F1 & TTFT (ms) & Total (ms) & Decode (tok/s) & Peak (GiB) \\
\midrule
Base & --- & 32.79 & 413.30 & 1011.22 & 829.19 & 14.942 \\
MLP Memory & FP32 & 33.68 & 420.77 & 1252.34 & 596.23 & 20.692 \\
MLP Memory & BF16 & 33.70 & 416.16 & 1099.91 & 725.03 & 17.817 \\
ResMem & FP32 & 34.77 & 420.79 & 1254.14 & 594.72 & 20.692 \\
ResMem & BF16 & 34.75 & 415.95 & 1097.76 & 726.77 & 17.817 \\
Memory Decoder & FP32 & 33.07 & 830.74 & 2029.42 & 413.20 & 17.921 \\
Memory Decoder & BF16 & 33.08 & 484.51 & 1658.11 & 422.42 & 15.955 \\
\bottomrule
\end{tabular}
\end{table}

\section{Tokenwise gate ablation}
\label{app:gate}
\begin{table}[!htbp]
\caption{Fixed-strength ResMem with and without the tokenwise gate. ResMem matches Table~\ref{tab:qa-main}.}
\label{tab:gate-ablation}
\centering
\footnotesize
\setlength{\tabcolsep}{3.5pt}
\begin{tabular}{lrrrr}
\toprule
 & \multicolumn{2}{c}{ResMem} & \multicolumn{2}{c}{ResMem + gate} \\
\cmidrule(lr){2-3}\cmidrule(lr){4-5}
Dataset & EM & F1 & EM & F1 \\
\midrule
NQ & 18.98 & 25.69 & 19.20 & 25.95 \\
WebQ & 17.32 & 31.94 & 17.13 & 31.65 \\
TriviaQA & 55.55 & 61.57 & 55.68 & 61.61 \\
Open-domain mean & 30.61 & 39.73 & 30.67 & 39.74 \\
\midrule
PopQA (long-tail) & 22.85 & 26.91 & 22.91 & 26.81 \\
HotpotQA (multi-hop) & 19.03 & 26.29 & 19.15 & 26.30 \\
\midrule
Five-dataset mean & 26.74 & 34.48 & 26.81 & 34.46 \\
\bottomrule
\end{tabular}
\end{table}

The gate decides when the memory should act on the base output.
The residual matters only at decoding steps where it changes the base's top token.
At such a step, it either repairs a base error or overrides a correct base token.
The gate aims to keep the repairs and block the overrides, using only signals available during decoding.

At step $t$, we compute the base logits $z_{B,t}$ and the fused logits $z_{B,t}+\resscale s_{\theta,t}$.
From both, we form 22 features.
They describe each distribution (entropy, top-token probability, and top-two margin) and how the two top tokens compare, such as the residual's advantage for the fused token.
They also cover the step index, the generated prefix, and whether either top token would end the answer.
A logistic classifier maps the features to a score $g_t$.
The step uses the fused logits if $g_t\geq 0.5$ and the base logits otherwise, so the residual scale is $\resscale_t\in\{0,\resscale\}$.
The gate uses no labels or future tokens.
On the test questions, the fused top token differs from the base top token at 19,254 of 463,520 decoding steps (4.2\%).
The gate accepts 10,614 of these changes and blocks 8,640.
Of the accepted changes, 7,493 occur at the first answer token.

The gate is fitted on 264 NQ/TriviaQA training questions and selected on a disjoint 4,091-question development split, so it adds QA supervision beyond memory training.
On all 38,627 questions, relative to fixed-strength ResMem, the gate changes mean EM by $+0.07$ points (95\% question-cluster bootstrap interval $[-0.06,0.19]$) and mean F1 by $-0.02$ points.
It alters 190 answers, 113 for the better and 77 for the worse, and the interval for every dataset includes zero.
Table~\ref{tab:qa-main} therefore reports ResMem without the gate.

\end{document}